\documentclass{article} % For LaTeX2e
\usepackage{iclr2027_conference,times}

\usepackage{amsmath,amsfonts,bm}

\def\eqref#1{equation~\ref{#1}}
\def\1{\bm{1}}

\DeclareMathAlphabet{\mathsfit}{\encodingdefault}{\sfdefault}{m}{sl}
\SetMathAlphabet{\mathsfit}{bold}{\encodingdefault}{\sfdefault}{bx}{n}

\usepackage{hyperref}
\usepackage{url}
\usepackage{booktabs}
\usepackage{graphicx}
\usepackage{multirow}
\usepackage[table]{xcolor}
\usepackage{wrapfig}
\usepackage{pifont}
\usepackage{titletoc}

\title{Thinking with Cameras: Active Visual Reasoning via Dynamic Viewpoint Control for Surveillance Video Understanding}

\author{
Xiao Zhang\textsuperscript{\rm 1,2}\quad
Wang Zeng\textsuperscript{\rm 2}\quad
Sheng Jin\textsuperscript{\rm 2}\thanks{Project Leader.}\quad
Wentao Liu\textsuperscript{\rm 2}\quad
Chen Qian\textsuperscript{\rm 2}\quad
Shichao Kan\textsuperscript{\rm 1}\thanks{Corresponding author.} \\
\textsuperscript{\rm 1}{School of Computer Science and Engineering, Central South University}\\
\textsuperscript{\rm 2}{SenseTime Research and Tetras.AI}\\
\texttt{xiaozhang@csu.edu.cn}\\
\texttt{\{zengwang, jinsheng, liuwentao, qianchen\}@tetras.ai}\\
\texttt{kanshichao@csu.edu.cn}
}

\iclrfinalcopy
\begin{document}

\maketitle
\lhead{}

\begin{abstract}
Large vision-language models (LVLMs) have recently achieved remarkable progress in general-purpose video understanding. However, their application to real-world surveillance remains challenging due to the lack of large-scale domain-specific datasets and the limitation of passive observation from fixed viewpoints. In surveillance scenarios, critical visual evidence can be easily missed when targets are distant, small, occluded, or move beyond the current camera view. In this work, we introduce CamVLM, a new framework for Thinking with Cameras, which enables LVLMs to actively acquire visual evidence in real-world surveillance by continuously controlling camera viewpoints. We first construct CCTV-Anomaly, a large-scale surveillance video understanding dataset containing 14,133 videos across 10 anomaly categories, with detailed captions and event annotations. We further formulate viewpoint control as an active visual perception problem and build CamTrack-53K, an object-centric viewpoint trajectory dataset for learning camera actions. Moreover, we propose a reinforcement learning based viewpoint policy optimization framework, which models camera control as a sequential decision-making process and learns long-horizon observation strategies beyond supervised trajectory imitation. Extensive experiments demonstrate that CamVLM achieves state-of-the-art performance under both passive observation and dynamic viewpoint settings, validating the effectiveness of active camera-based reasoning for surveillance video understanding. Our datasets, model, and code will be available at \url{https://github.com/xiaozhang79/CamVLM}.
\end{abstract}

\section{Introduction}
\label{sec:intro}

Surveillance cameras are widely deployed in real-world applications, including anomaly detection, behavior analysis, and traffic perception. Unlike general videos collected for human viewing, streaming surveillance videos usually cover wide fields of view and contain multiple moving and interacting objects. Due to long capture distances, target objects often occupy only small regions of the entire frame and may gradually move beyond the current camera viewpoint, leading to incomplete and unreliable visual evidence. Therefore, effective surveillance video understanding requires not only recognizing what happens in a video, but also actively determining where to observe next by continuously adjusting the camera viewpoint to acquire informative visual evidence.

\begin{wrapfigure}{r}{0.5\textwidth}
\vspace{-0.15in}
\centering
\includegraphics[width=\linewidth]{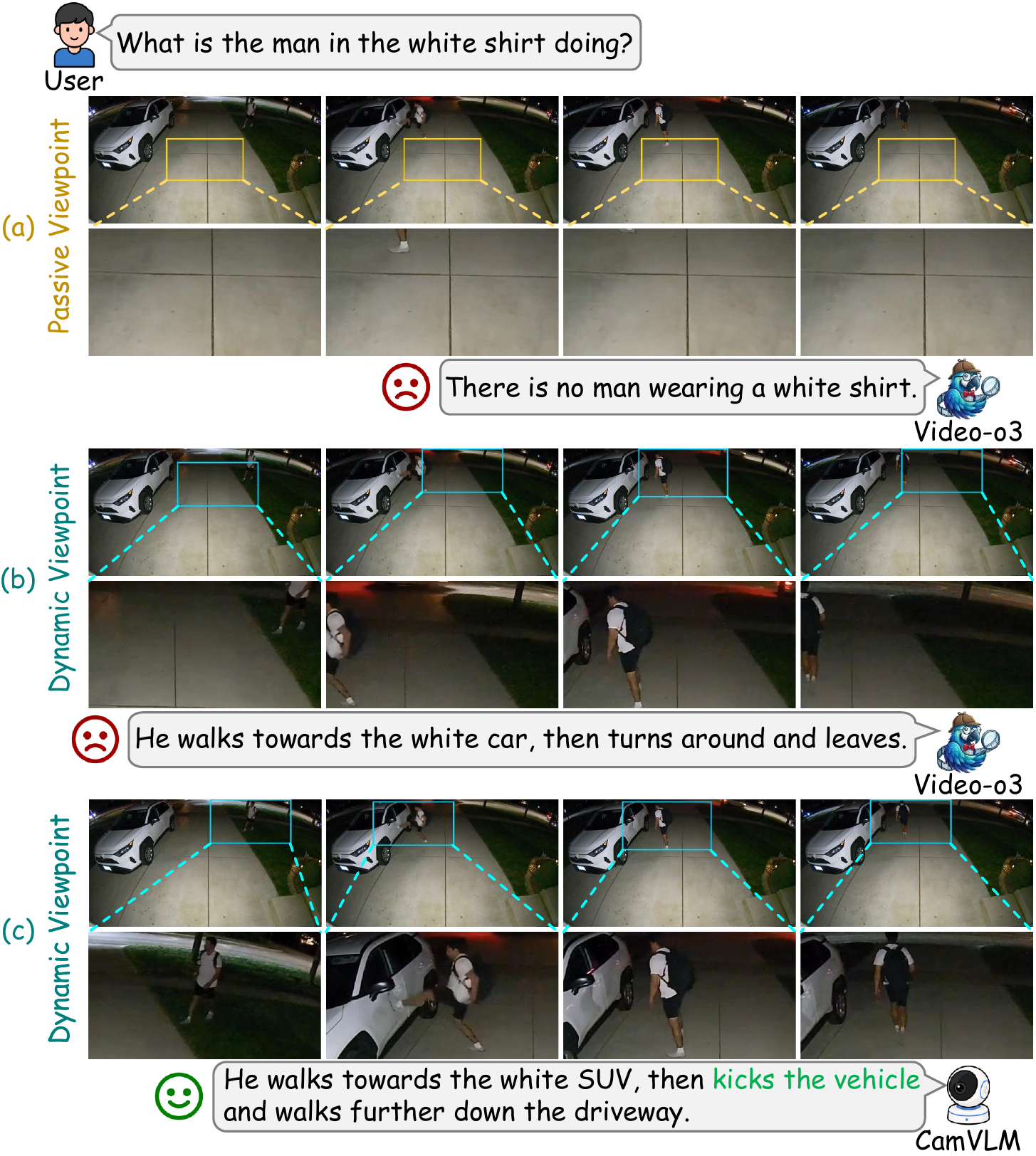}
\caption{Motivation of CamVLM. The full frame represents the virtual panoramic space, while the bounding box denotes the camera viewpoint. (a) Passive viewpoints may miss critical cues. (b,c) Under dynamic viewpoint control, existing methods such as Video-o3~\citep{zeng2026video} exhibit unstable visual focus, whereas CamVLM reliably captures fine-grained visual evidence.}
\label{fig:motivation}
\vspace{-0.15in}
\end{wrapfigure}

Recent advances in large vision-language models (LVLMs) have significantly improved visual perception and reasoning. Early video understanding approaches~\citep{feng2025video, li2025videochat, wang2025videorft} typically encode visual contents globally and perform subsequent language reasoning, which often overlooks fine-grained visual evidence. Inspired by OpenAI-o3~\citep{openai2025o3}, recent ``thinking with images'' approaches~\citep{khayatkhoei2025mllms, zheng2026deepeyes, shen2025zoomeye} introduce iterative evidence acquisition, where models actively retrieve informative regions and interleave visual perception with reasoning. Extending this idea, recent ``thinking with videos'' methods~\citep{yang2026longvt, zhang2026thinking, zeng2026video} further explore dynamic visual evidence acquisition from videos through iterative frame sampling or tool-based interaction. However, existing thinking-based approaches still operate on fixed visual streams. They can decide which frame or clip to inspect, but cannot determine how the camera itself should move to acquire better observations. This limitation is particularly critical in real-world surveillance scenarios, where objects are often small, partially occluded, or follow unpredictable trajectories. As shown in Figure~\ref{fig:motivation}(a), once important objects leave the current viewpoint, passive video retrieval strategies cannot recover the missing fine-grained visual evidence.

\begin{figure}[h]
\centering
\includegraphics[width=\textwidth]{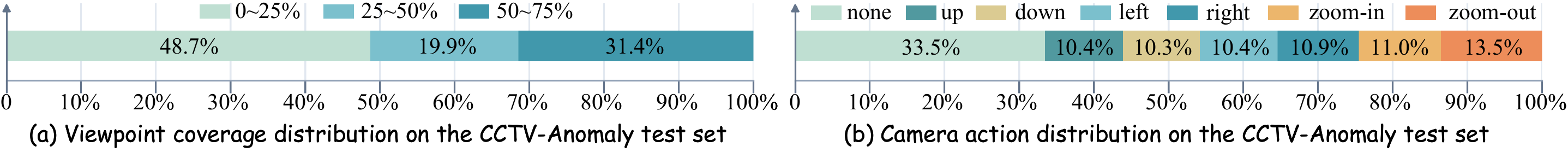}
\caption{Behavior statistics of the learned camera control policy on the CCTV-Anomaly test set. (a) Distribution of selected viewpoint coverage. Nearly half of the viewpoints remain within 25\% of the full frame, indicating that the model avoids unnecessary zoom-out and adaptively adjusts the viewing scale according to scene context. (b) Distribution of camera control actions. The high proportion of none actions and the balanced directional movements and zooming operations demonstrate an efficient on-demand control strategy without directional or zooming bias.}
\label{fig:distribution}
\end{figure}

To address this challenge, we propose Thinking with Cameras, a new active visual reasoning paradigm that enables LVLMs to continuously reason through camera actions in real-world surveillance. However, training LVLMs to control real-world cameras is hindered by the lack of large-scale viewpoint-action trajectories collected from physical surveillance systems. To overcome this limitation, we introduce a virtual camera formulation by treating each full video frame as a panoramic observation space and a local region as the camera's current field of view. Specifically, the model starts from an initial viewpoint and learns to perform camera-like operations, including panning and zooming, to acquire informative visual evidence. Unlike treating surveillance videos as immutable visual streams, our formulation transforms visual understanding into an active perception process, where the model dynamically determines where to observe next. Unlike conventional visual reasoning, dynamic viewpoint control is inherently a sequential decision-making problem, since each camera action changes future observations and influences subsequent reasoning outcomes. Therefore, we further introduce a viewpoint policy optimization framework based on reinforcement learning, which enables LVLMs to learn long-horizon camera control strategies beyond supervised trajectory imitation. By jointly optimizing task completion and visual evidence acquisition, the learned policy can explore more effective viewpoint trajectories and continuously maintain informative objects within the field of view. As shown in Figure~\ref{fig:distribution}, the learned camera control policy exhibits adaptive viewpoint selection and on-demand camera actions, suggesting that efficient surveillance understanding benefits from active and evidence-driven viewpoint reasoning.

To support this paradigm, we construct two large-scale datasets. First, we introduce CCTV-Anomaly, a high-quality surveillance video understanding dataset containing 14,133 videos across 10 anomaly categories and 234 hours, with detailed captions and event category annotations. Second, we build CamTrack-53K, an object-centric viewpoint trajectory dataset for learning dynamic camera control. Based on these datasets and the proposed viewpoint policy optimization framework, we develop CamVLM, which actively controls camera viewpoints during reasoning and captures comprehensive fine-grained object-level spatiotemporal evidence, as illustrated in Figure~\ref{fig:motivation}(b,c).

Overall, our contributions are summarized as follows:
\begin{itemize}
\item We propose Thinking with Cameras, a new active visual reasoning paradigm that enables LVLMs to continuously control camera viewpoints and actively acquire visual evidence in real-world surveillance.

\item We formulate dynamic viewpoint control as a sequential decision-making problem and introduce a reinforcement learning based viewpoint policy optimization framework, which learns long-horizon camera control strategies by jointly optimizing reasoning accuracy and visual evidence acquisition.

\item We construct CCTV-Anomaly and CamTrack-53K, two large-scale datasets for surveillance video understanding and dynamic viewpoint control, respectively.

\item We develop CamVLM, an LVLM with dynamic viewpoint control capability that actively determines where to observe next and achieves more reliable fine-grained surveillance video understanding.
\end{itemize}
\section{Related Work}
\label{sec:related}

\paragraph{Thinking with Images and Videos.}
Recent advances in LVLM reasoning have explored iterative evidence acquisition beyond conventional text-centric reasoning. Inspired by OpenAI-o3~\citep{openai2025o3}, ``thinking with images'' methods~\citep{lai2026mini, khayatkhoei2025mllms, zheng2026deepeyes, shen2025zoomeye} interleave visual perception with language reasoning by repeatedly retrieving fine-grained image evidence, improving detailed visual understanding. Extending this idea to videos, recent ``thinking with videos'' approaches~\citep{fu2025love, yang2026longvt, zhang2026thinking, zeng2026video} enable LVLMs to iteratively access informative frames or clips through tool use and dynamic perception. However, these methods still operate on fixed visual streams and focus on selecting existing visual evidence rather than actively controlling the camera viewpoint, making them insufficient for real-world surveillance scenarios where targets may be distant, occluded, or leave the current field of view.

\paragraph{Surveillance Video Datasets.}
Existing surveillance video datasets mainly focus on anomaly detection, such as Subway Entrance, Subway Exit~\citep{adam2008robust}, UCSDPed1/2~\citep{li2014anomaly}, Avenue~\citep{lu2013abnormal}, and UCF-Crime~\citep{sultani2018real}. However, these datasets are limited in scale, realism, or annotation richness, providing insufficient supervision for modern LVLM-based surveillance understanding. Recent UDVideoQA~\citep{vishal2026udvideoqa} explores surveillance video question answering in traffic scenarios but contains limited video duration and domain coverage. In contrast, our CCTV-Anomaly provides 14,133 high-quality surveillance videos across diverse anomaly categories, with both detailed captions and event-level annotations to support comprehensive surveillance video understanding.

\paragraph{Active Visual Perception.}
Active perception aims to improve visual understanding by dynamically adjusting sensing strategies according to task requirements. Existing works mainly focus on robotic perception~\citep{zhang2025instance, liu2026activevla}, object tracking~\citep{wang2025trackvla, sun2025open}, and embodied agents~\citep{qin2024mp5, liu2025activevoo}, where sensors or cameras are controlled through predefined policies or reinforcement learning strategies. However, applying active perception to LVLM-based surveillance understanding remains largely unexplored. Different from existing approaches that passively analyze captured videos, we introduce a ``thinking with cameras'' paradigm, where LVLMs actively control viewpoints and learn long-horizon camera policies through reinforcement learning to acquire task-relevant visual cues.
\section{Method}
\label{sec:method}

\begin{figure*}[t]
\centering
\includegraphics[width=\textwidth]{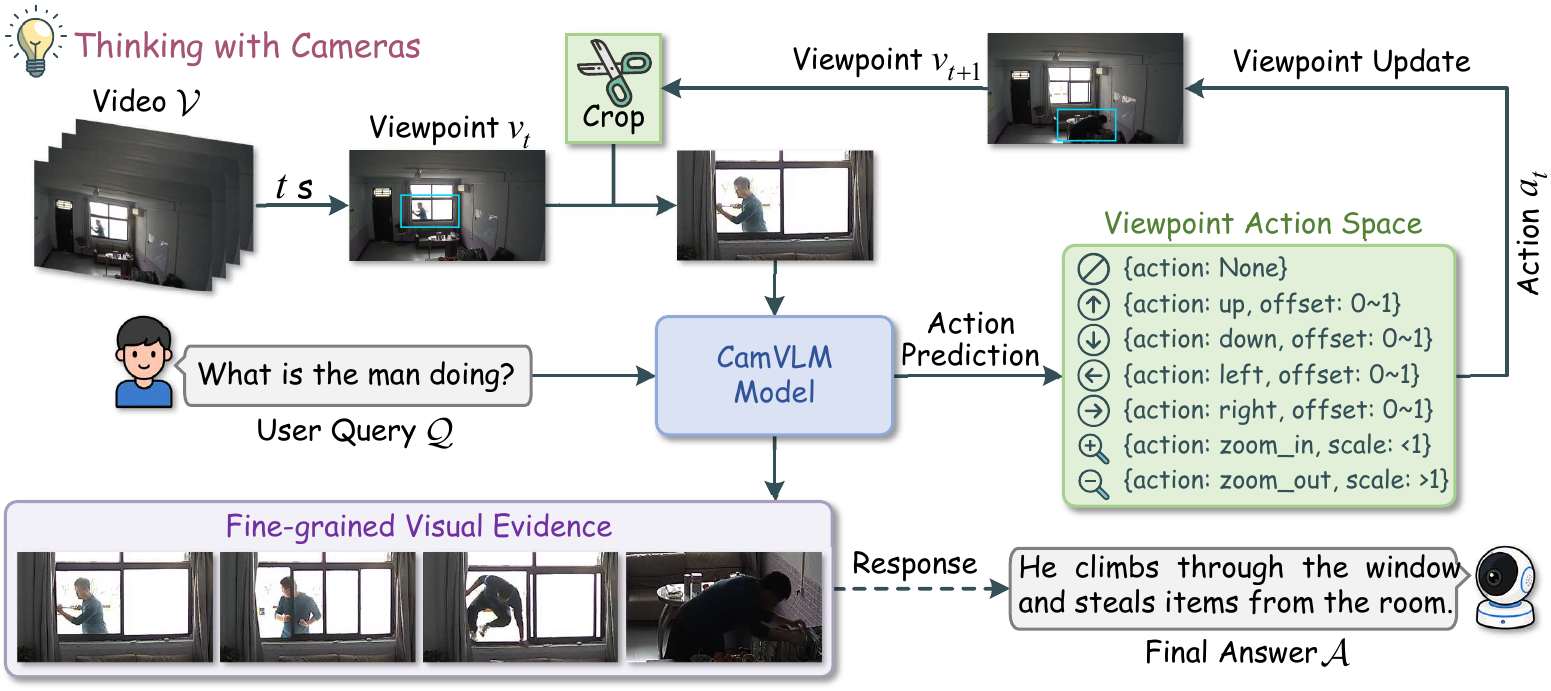}
\caption{Overview of CamVLM. Given a surveillance video and a user query, CamVLM iteratively predicts viewpoint actions based on the positions of task-relevant objects, updates the viewpoint to keep them clearly visible, accumulates fine-grained visual evidence, and produces the final answer.}
\label{fig:overview}
\end{figure*}

\subsection{Overview}
\label{sec:overview}

As illustrated in Figure~\ref{fig:overview}, we propose CamVLM, a vision-language model that actively controls camera viewpoints to acquire visual evidence in real-world surveillance. Different from conventional LVLMs that passively analyze fixed visual streams, CamVLM treats viewpoint adjustment as an integral part of the reasoning process and actively acquires informative visual evidence through camera actions.
Given a video $\mathcal{V}$ and a user query $\mathcal{Q}$, the model maintains a dynamic viewpoint state $v_t$ at each timestep $t$. Based on the current visual observation and task requirements, CamVLM predicts a viewpoint action $a_t$ to determine how the camera should move or zoom. The executed action updates the viewpoint and provides a new observation $v_{t+1}$, enabling the model to progressively collect fine-grained evidence. After multiple interaction steps, the accumulated visual evidence is integrated to generate the final response $\mathcal{A}$. Formally, the viewpoint interaction process can be represented as:
\begin{equation}
a_t=\pi_\theta(\mathcal{Q},v_t),
\qquad
v_{t+1}=\mathcal{T}(v_t,a_t),
\end{equation}
where $\pi_\theta$ denotes the viewpoint policy and $\mathcal{T}$ represents the viewpoint transformation operation. The complete interaction trajectory is defined as:
\begin{equation}
\tau=
(\mathcal{Q},v_0,a_0,v_1,a_1,\ldots,v_{T-1},a_{T-1},\mathcal{A}).
\end{equation}

The following subsections introduce the proposed dynamic viewpoint control mechanism, the construction of CamTrack-53K for learning viewpoint policies, and the training strategy including reinforcement learning based viewpoint policy optimization.

\subsection{Dynamic Viewpoint Control}
\label{sec:approach}

Dynamic viewpoint control aims to enable LVLMs to actively determine where to observe next in real-world surveillance in order to acquire task-relevant visual evidence. Ideally, such capability requires large-scale interaction data collected from physical cameras operating in real-world environments, which is difficult to obtain. Therefore, we simulate camera control within surveillance videos by treating each full-frame image as a virtual panoramic space and a local region as the current camera observation. For each input video frame, the central $1/9$ region is initially selected as the camera viewpoint $v_0$. This setting provides a suitable balance between observation coverage and control necessity: larger viewpoints reduce the need for active adjustment, while smaller ones may miss sufficient target information.

During inference, the model interacts with the video at one-second intervals. At timestep $t$, given the current viewpoint $v_t$ and query $\mathcal{Q}$, the viewpoint policy predicts an action $a_t$ according to the spatial state of task-relevant objects and the current visual evidence. The action space contains both spatial movement and scale adjustment operations, allowing the model to actively explore the surrounding visual space. After executing the predicted action, the viewpoint is updated through the transformation function:
\begin{equation}
v_{t+1}=\mathcal{T}(v_t,a_t).
\end{equation}
Through iterative viewpoint adjustment, CamVLM can continuously acquire complementary observations in real-world surveillance when objects become distant, partially visible, or leave the initial field of view. Unlike passive video sampling methods, the proposed mechanism enables the model to actively decide which visual evidence to acquire before producing the final answer.

\subsection{CamTrack-53K: Viewpoint Trajectory Generation}
\label{sec:camtrack}
Training dynamic viewpoint policies requires large-scale supervision of camera-object interactions. To this end, we construct CamTrack-53K, an object-centric viewpoint trajectory dataset for learning active camera control. Specifically, we collect videos from A2D~\citep{xu2015can}, MeViS~\citep{ding2023mevis}, and Refer-YouTube-VOS~\citep{seo2020urvos} in VideoRefer-700K~\citep{yuan2025videorefer}. These datasets provide temporally aligned object masks and object-level question-answer annotations, enabling the construction of task-oriented viewpoint trajectories. For each video, object masks are first converted into bounding boxes. The viewpoint action space is defined as:
 \(\{\texttt{none},
\texttt{up},\texttt{down},\texttt{left},\texttt{right},\texttt{zoom\_in},
\texttt{zoom\_out}\}\).

Panning actions are parameterized by offsets relative to the full frame, while zooming actions are represented by scale changes relative to the current viewpoint. Starting from the initial central $1/9$ viewpoint, we generate trajectories at one-second intervals based on object-viewpoint relationships:

\begin{itemize}
\item When the target object is outside the current viewpoint, the camera performs zoom-out and viewpoint adjustment actions to reacquire the object.
\item When the object is visible but deviates from the optimal observation region, the camera performs panning and zooming actions to obtain a more informative view.
\item When the object is already well observed, the camera maintains the current viewpoint.
\end{itemize}

The generated trajectories provide an initial policy prior for supervised learning, while subsequent reinforcement learning further refines the viewpoint policy beyond predefined trajectory patterns.

\subsection{Training Pipeline}
\label{sec:training}

\paragraph{Supervised Fine-Tuning.}
We adopt a mixed supervised fine-tuning (SFT) strategy on the constructed CCTV-Anomaly and CamTrack-53K datasets to jointly enhance the model's surveillance video understanding and dynamic viewpoint control capabilities. For CCTV-Anomaly, the model is trained under the passive full-frame observation setting, where only the final event classification and video captioning responses are supervised. For CamTrack-53K, we follow the dynamic viewpoint interaction process described in Section~\ref{sec:approach}. Specifically, the model is supervised to predict both the viewpoint control actions and the final reasoning response according to the generated viewpoint trajectories. The action space follows the definition in Section~\ref{sec:camtrack}. When multiple viewpoint adjustments are required, the model predicts an ordered action sequence, such as \(\{\texttt{action}: \texttt{up}, \texttt{offset}: \texttt{0.1}; \texttt{action}: \texttt{zoom\_in}, \texttt{scale}: \texttt{0.8}\}\).

\paragraph{Viewpoint Policy Optimization.}
Dynamic viewpoint control is inherently a sequential decision-making problem. Each camera action determines the visual observations available at subsequent reasoning steps, making its quality dependent on long-term visual evidence acquisition rather than immediate supervision. Although supervised fine-tuning enables the model to imitate expert viewpoint trajectories, it cannot optimize viewpoint policies for downstream understanding, since multiple camera trajectories may lead to equally informative observations. To address this limitation, we further optimize CamVLM on CamTrack-53K using reinforcement learning (RL) with Group Relative Policy Optimization (GRPO)~\citep{shao2024deepseekmath}, allowing the model to actively explore more effective viewpoint control strategies.

Unlike conventional reinforcement learning for large language models, where each sampled response consists only of textual outputs, each response in CamVLM contains an entire camera-control trajectory together with the final reasoning result. Therefore, policy optimization jointly evaluates both visual exploration behaviors and downstream reasoning performance. Specifically, given an input video $\mathcal{V}$ and a user query $\mathcal{Q}$, we follow the interaction process described in Section~\ref{sec:approach} to generate a complete camera-control trajectory
\begin{equation}
\tau=\{a_1,a_2,\cdots,a_T,y\},
\end{equation}
where $a_t$ denotes the viewpoint action at time step $t$, and $y$ is the final answer.

For each training sample $(\mathcal{V},\mathcal{Q})\in\mathcal{D}$, we sample a group of $G$ trajectories
$\{o_1,o_2,\cdots,o_G\}$ from the policy model $\pi_\theta$, compute their corresponding rewards
$\{r_1,r_2,\cdots,r_G\}$, and optimize the policy using GRPO:
\begin{equation}
\mathcal{L}_{\mathrm{GRPO}}
=
-\mathbb{E}_{(\mathcal{V},\mathcal{Q})\sim\mathcal{D}}
\mathbb{E}_{o_i\sim\pi_\theta}
\left[
A_i
\log
\pi_\theta
(o_i|\mathcal{V},\mathcal{Q})
\right],
\end{equation}
where the relative advantage is computed as
\begin{equation}
A_i
=
\frac{
r_i-\mathrm{mean}(\{r_1,\cdots,r_G\})
}{
\mathrm{std}(\{r_1,\cdots,r_G\})
}.
\end{equation}

To jointly optimize surveillance understanding and active visual perception, we design a task-oriented reward composed of a task completion reward and a visual evidence reward:
\begin{equation}
\mathcal{R}
=
\mathcal{R}_{\mathrm{task}}
+
\lambda
\mathcal{R}_{\mathrm{visual}},
\end{equation}
where $\lambda$ balances the two objectives. The task completion reward evaluates the correctness of the final result. For multiple-choice question answering, a correct prediction receives $1$, while an incorrect prediction receives $0$:
\begin{equation}
\mathcal{R}_{\mathrm{task}}
=
\begin{cases}
1,
&
\text{if the predicted answer is correct},
\\
0,
&
\text{otherwise}.
\end{cases}
\end{equation}
The visual evidence reward evaluates the quality of the observations acquired by the predicted camera policy. Instead of directly encouraging imitation of reference trajectories, it measures how well the predicted viewpoints continuously cover informative target objects throughout the interaction:
\begin{equation}
\mathcal{R}_{\mathrm{visual}}
=
\frac{1}{T}
\frac{1}{N}
\sum_{t=0}^{T-1}
\sum_{i=1}^{N}
\frac{
|v_{t+1}\cap b_{t+1}^{(i)}|
}{
|b_{t+1}^{(i)}|
},
\end{equation}
where $T$ is the number of interaction steps, $N$ is the number of target objects, $v_{t+1}$ is the predicted camera viewpoint at the next step, and $b_{t+1}^{(i)}$ is the ground-truth bounding box of the $i$-th object.

By rewarding observation quality rather than exact camera trajectories, the proposed policy optimization allows multiple valid viewpoint strategies to emerge during training. Consequently, CamVLM learns to actively determine where to observe next, continuously maintain informative objects within the field of view, and acquire richer visual evidence for downstream reasoning.
\section{CCTV-Anomaly Dataset}
\label{sec:dataset}

\begin{wraptable}{r}{0.5\textwidth}
\vspace{-0.3in}
\centering
\caption{Comparison of CCTV-Anomaly with existing surveillance anomaly video datasets.}
\label{tab:dataset_comparison}
\resizebox{\linewidth}{!}{%
\begin{tabular}{lccc}
\toprule
\textbf{Dataset} & \textbf{\#Videos} & \textbf{Duration} & \textbf{Caption} \\
\midrule
Subway Entrance & 1 & 1.5 h & \ding{55} \\
Subway Exit & 1 & 1.5 h & \ding{55} \\
UCSD Ped1 & 70 & 5 min & \ding{55} \\
UCSD Ped2 & 28 & 5 min & \ding{55} \\
Avenue & 37 & 30 min & \ding{55} \\
UCF-Crime & 1,900 & 128 h & \ding{55} \\
\rowcolor[RGB]{216,229,223}
\textbf{CCTV-Anomaly}
& \textbf{14,133}
& \textbf{234 h}
& \ding{51} \\
\bottomrule
\end{tabular}%
}
\vspace{-0.15in}
\end{wraptable}

\begin{figure*}[!t]
\centering
\includegraphics[width=\textwidth]{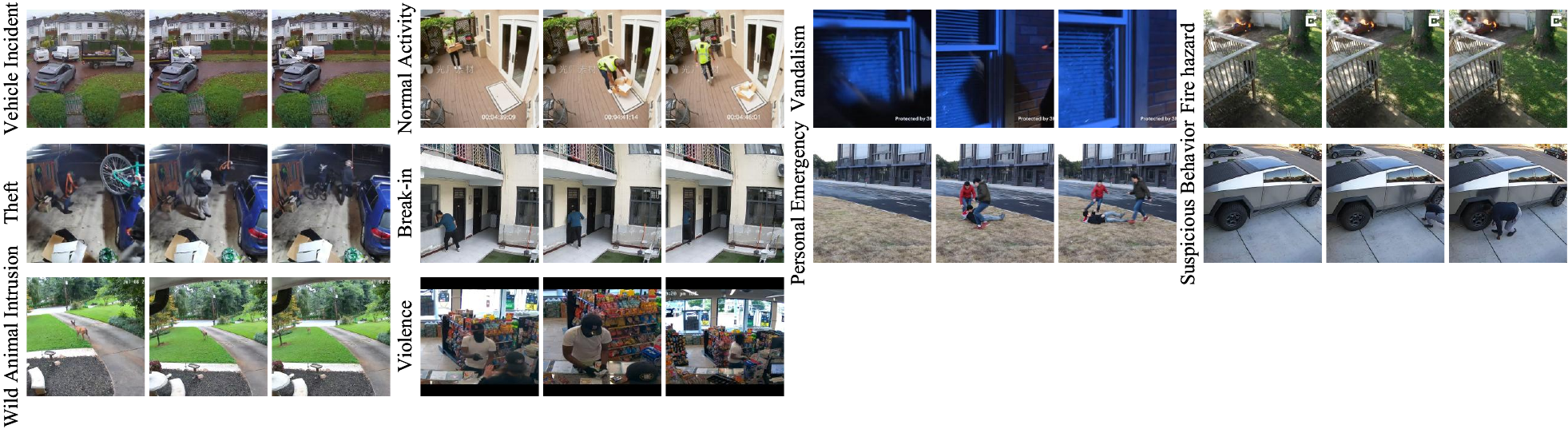}
\caption{Examples of different anomalies in our CCTV-Anomaly dataset.}
\label{fig:data_examples}
\end{figure*}

Existing surveillance anomaly datasets mainly provide coarse event labels, which are insufficient for fine-grained LVLM reasoning. To address this limitation, we construct CCTV-Anomaly, a large-scale surveillance video understanding dataset containing 14,133 real-world videos across 10 event categories. Each video is annotated with both an event category and a detailed natural language caption, enabling joint evaluation of event classification and video captioning. Table~\ref{tab:dataset_comparison} compares CCTV-Anomaly with existing datasets, and Figure~\ref{fig:data_examples} presents representative examples.

\paragraph{Dataset Construction and Split.}

CCTV-Anomaly is built from both existing public surveillance datasets and newly collected real-world surveillance videos, covering diverse environments and viewpoints. Each video is annotated using an LLM-assisted pipeline followed by human verification. In addition to event categories and captions, we further annotate the test set with structured semantic labels, including location, time of day, subjects, activities, and objects, to support fine-grained evaluation. More details are provided in the Appendix~\ref{sec:sup_cctv}.

The dataset is divided into 13,133 training videos and 1,000 testing videos, with 100 testing videos for each event category. CCTV-Anomaly supports two benchmark tasks: event classification and video captioning. Detailed evaluation protocols are described in Section~\ref{sec:experiments}.
\section{Experiments}
\label{sec:experiments}

\subsection{Experimental Setup}
\paragraph{Implementation Details.}
We adopt Qwen3-VL-8B-Instruct~\citep{bai2025qwen3} as the base LVLM. During training and evaluation, videos are uniformly sampled at 2 FPS, with a maximum of 14,336 video tokens and 448 frames. The visual encoder is frozen, while all remaining parameters are optimized. For supervised fine-tuning, we use a global batch size of 128 with a learning rate of \(1\times10^{-5}\). For reinforcement learning, we employ GRPO-based viewpoint policy optimization with a batch size of 8 and 8 sampled trajectories per prompt. The learning rate is set to \(1\times10^{-6}\), the KL coefficient \(\beta\) is set to 0, and the viewpoint alignment reward weight \(\lambda\) is set to 0.6. All experiments are conducted on 8 NVIDIA A800 GPUs.

\paragraph{Benchmarks and Evaluation Metrics.}
We evaluate surveillance video understanding on two benchmarks: CCTV-Anomaly and UDVideoQA~\citep{vishal2026udvideoqa}. For CCTV-Anomaly, we report event classification accuracy and video captioning accuracy. Classification accuracy is obtained by extracting the predicted category using regular expressions and comparing it with the ground-truth label. Captioning accuracy is evaluated over five semantic components (location, time of day, subjects, activities, and objects), each contributing 20 points to a total score of 100. We use GPT-5.5~\citep{openai2026gpt55} to semantically evaluate whether the generated caption includes each ground-truth component, followed by manual verification, and report the average score over the test set. For UDVideoQA, since the official test split is unavailable, we construct UDVideoQA$^*$ with permission from the original authors using Set~03 and Set~20 from the released dataset, which provide the greatest question diversity. The resulting benchmark contains 1,190 QA pairs.
Following the official evaluation protocol, we report accuracy across five question types and the weighted overall accuracy, where more complex reasoning types receive larger weights. The weights are 1.0 for Basic Understanding (BU), 1.2 for Attribution (Atr), 1.3 for Event Reasoning (ER) and Reverse Reasoning (RR), and 1.5 for Counterfactual Inference (CI).

\paragraph{Evaluation Settings.}
To evaluate the effectiveness of active camera reasoning, we consider two complementary settings: passive viewpoint understanding and dynamic viewpoint control.

\textbf{Passive Viewpoint Setting.}
The model receives the complete video frames or a fixed viewpoint without any camera adjustment. This setting evaluates conventional surveillance video understanding under passive observation.

\textbf{Dynamic Viewpoint Setting.}
The model starts from the central \(1/9\) region of each frame and actively adjusts the viewpoint through predicted camera actions. Only the visual regions acquired through the dynamic viewpoint trajectory are provided for final answer generation. This setting evaluates whether the model can actively acquire informative visual evidence and improve surveillance video understanding through camera control.

\begin{wraptable}{r}{0.5\textwidth}
\vspace{-0.3in}
\centering
\caption{Performance comparisons on CCTV-Anomaly under the conventional full-frame passive observation setting. The best results are in bold.}
\label{tab:main_result_1_1}
\resizebox{\linewidth}{!}{%
\begin{tabular}{lc|cc}
\toprule
\multirow{2}{*}{\textbf{Methods}}
& \multirow{2}{*}{\textbf{Sizes}}
& \multicolumn{2}{c}{\textbf{CCTV-Anomaly}} \\
\cmidrule(lr){3-4}
& & Cls Avg & Cap Avg \\
\midrule
Gemini-3.1-Flash-Lite-Preview
& \multirow{2}{*}{-}
& \multirow{2}{*}{59.3}
& \multirow{2}{*}{63.0} \\
\citep{google2026gemini31flashlite}
& & & \\
GPT-5.6-Luna~\citep{openai2026gpt56}
& - & 57.6 & 61.8 \\
Doubao-Seed-2.0-Lite~\citep{seed2026seed2}
& - & 56.5 & 65.4 \\
\midrule
Video-R1~\citep{feng2025video}
& 7B & 43.9 & 54.0 \\
VideoZoomer~\citep{ding2026videozoomer}
& 7B & 49.0 & 53.2 \\
MiMo-VL~\citep{yue2025mimo}
& 7B & 52.8 & 57.5 \\
Video-o3~\citep{zeng2026video}
& 7B & 53.2 & 60.0 \\
LLaVA-OV-2~\citep{an2026llava}
& 8B & 45.3 & 66.6 \\
Qwen3-VL~\citep{bai2025qwen3}
& 8B & 46.0 & 61.8 \\
\rowcolor[RGB]{216,229,223}
\textbf{CamVLM (Ours)}
& 8B & \textbf{75.6} & \textbf{76.3} \\
\bottomrule
\end{tabular}%
}
\vspace{-0.15in}
\end{wraptable}

\subsection{Experiment Results}

\paragraph{Results on CCTV-Anomaly.}
Table~\ref{tab:main_result_1_1} compares surveillance video understanding on CCTV-Anomaly under full-frame observation. CamVLM achieves the best performance on both event classification and video captioning, demonstrating strong surveillance understanding capability. Under dynamic viewpoints, CamVLM outperforms passive-viewpoint methods on both tasks (Table~\ref{tab:main_result_1_9}), highlighting the effectiveness of active viewpoint adjustment. Compared with the passive baseline, dynamic viewpoint control improves event classification by \(33.8\%\), while CamVLM achieves further gains by capturing fine-grained object-centric spatiotemporal evidence.

\begin{table*}[ht]
\caption{Performance comparisons on CCTV-Anomaly and UDVideoQA\(^*\) under passive and dynamic viewpoint settings. The terms ``Passive'' and ``Dynamic'' denote a passive viewpoint fixed to the central \(1/9\) region and a dynamic viewpoint initialized from the same region and adjusted over time, respectively. The best results are in bold.}
\label{tab:main_result_1_9}
\centering
\small
\setlength{\tabcolsep}{2.35pt}
\begin{tabular}{lcc|cc|cccccc}
\toprule
& & & \multicolumn{2}{c|}{\textbf{CCTV-Anomaly}} & \multicolumn{6}{c}{\textbf{UDVideoQA\(^*\)}} \\
\cmidrule(lr){4-5}
\cmidrule(lr){6-11}
\multirow{-2}{*}{\textbf{Methods}}
& \multirow{-2}{*}{\textbf{Sizes}}
& \multirow{-2}{*}{\textbf{Settings}}
& Cls Avg
& Cap Avg
& BU & Atr & ER & RR & CI & Avg \\
\midrule
Gemini-3.1-Flash-Lite-Preview
& \multirow{2}{*}{-}
& \multirow{2}{*}{Passive}
& \multirow{2}{*}{45.0}
& \multirow{2}{*}{52.3}
& \multirow{2}{*}{61.3}
& \multirow{2}{*}{17.6}
& \multirow{2}{*}{44.1}
& \multirow{2}{*}{67.2}
& \multirow{2}{*}{79.0}
& \multirow{2}{*}{54.9} \\
\citep{google2026gemini31flashlite}
& & & & & & & & & & \\
GPT-5.6-Luna~\citep{openai2026gpt56} & - & Passive & 42.4 & 51.4 & 56.3 & 13.9 & 38.7 & 59.7 & 32.8 & 39.7 \\
Doubao-Seed-2.0-Lite~\citep{seed2026seed2} & - & Passive & 37.2 & 51.0 & 55.9 & 21.0 & 48.3 & 51.3 & 66.8 & 49.3 \\
\midrule
Video-R1~\citep{feng2025video} & 7B & Passive & 37.0 & 46.1 & 52.5 & 11.8 & 36.6 & 41.2 & 69.8 & 43.3 \\
VideoZoomer~\citep{ding2026videozoomer} & 7B & Passive & 37.8 & 46.7 & 56.3 & 10.1 & 35.3 & 29.0 & 63.9 & 39.3 \\
MiMo-VL~\citep{yue2025mimo} & 7B & Passive & 38.8 & 47.1 & 55.5 & 10.5 & 39.9 & 52.9 & 51.3 & 42.2 \\
Video-o3~\citep{zeng2026video} & 7B & Passive & 41.4 & 50.2 & 56.7 & 15.5 & 42.0 & 39.9 & 56.3 & 42.3 \\
LLaVA-OneVision-2~\citep{an2026llava} & 8B & Passive & 37.3 & 55.6 & 57.6 & 14.3 & 40.8 & 55.9 & 45.0 & 42.5 \\
Qwen3-VL~\citep{bai2025qwen3} & 8B & Passive & 36.7 & 51.3 & 57.1 & 10.5 & 41.2 & 46.2 & 51.3 & 41.3 \\
Qwen3-VL~\citep{bai2025qwen3} & 8B & Dynamic & 44.0 & 57.2 & 67.2 & 17.2 & 44.5 & 51.3 & 51.3 & 45.9 \\
\rowcolor[RGB]{216,229,223}
\textbf{CamVLM (Ours)} & 8B & Dynamic & \textbf{70.5} & \textbf{69.7} & \textbf{70.6} & \textbf{29.4} & \textbf{46.2} & \textbf{64.7} & \textbf{72.7} & \textbf{57.0} \\
\bottomrule
\end{tabular}
\end{table*}

\paragraph{Generalization Results on UDVideoQA\(^*\).}
Table~\ref{tab:main_result_1_9} reports zero-shot results on UDVideoQA\(^*\). CamVLM achieves the best performance across all question types under dynamic viewpoints. Compared with passive observation, dynamic viewpoint control improves the baseline by \(15.7\%\) overall, demonstrating strong generalization to unseen surveillance scenarios.

\subsection{Ablation Studies}
To validate the effectiveness of our design, we conduct comprehensive ablation studies on UDVideoQA\(^*\). Unless otherwise specified, passive viewpoints are fixed to the central \(1/9\) region, while dynamic viewpoints are initialized from the same region. All models use the 8B configuration.

\begin{wraptable}{r}{0.5\textwidth}
\vspace{-0.15in}
\centering
\caption{Ablation study of dynamic viewpoint control on UDVideoQA\(^*\). The best results are in bold. The default configuration is in gray.}
\label{tab:ablation_viewpoint}
\resizebox{\linewidth}{!}{%
\begin{tabular}{lc|cccccc}
\toprule
& & \multicolumn{6}{c}{\textbf{UDVideoQA\(^*\)}} \\
\cmidrule(lr){3-8}
\multirow{-2}{*}{\textbf{Methods}}
& \multirow{-2}{*}{\textbf{Settings}}
& BU & Atr & ER & RR & CI & Avg \\
\midrule
Qwen3-VL & Passive
& 57.1 & 10.5 & 41.2 & 46.2 & 51.3 & 41.3 \\
Qwen3-VL & Dynamic
& 67.2 & 17.2 & 44.5 & 51.3 & 51.3 & 45.9 \\
CamVLM & Passive
& 59.2 & 22.3 & 41.6 & 52.1 & 60.4 & 47.4 \\
\rowcolor[RGB]{235,235,235}
CamVLM & Dynamic
& \textbf{70.6}
& \textbf{29.4}
& \textbf{46.2}
& \textbf{64.7}
& \textbf{72.7}
& \textbf{57.0} \\
\bottomrule
\end{tabular}%
}
\vspace{-0.15in}
\end{wraptable}

\paragraph{Effect of Dynamic Viewpoint Control.}
Table~\ref{tab:ablation_viewpoint} compares passive and dynamic viewpoint settings. Dynamic viewpoints improve both the baseline model and CamVLM, with substantially larger gains for CamVLM. This demonstrates that viewpoint control effectively directs the model to informative regions, while CamVLM better exploits the resulting object-centric spatiotemporal evidence. We further evaluate CamVLM with viewpoints fixed to the central \(1/9\) region during inference. The significant performance drop confirms that the improvements mainly stem from effective viewpoint control rather than the video understanding capability learned during training.

\begin{wraptable}{r}{0.5\textwidth}
\vspace{-0.3in}
\centering
\caption{Ablation study of different viewpoint sizes on UDVideoQA\(^*\). The best results are in bold. The default configuration is in gray.}
\label{tab:ablation_viewpoint_size}
\resizebox{\linewidth}{!}{%
\begin{tabular}{lc|cccccc}
\toprule
& & \multicolumn{6}{c}{\textbf{UDVideoQA\(^*\)}} \\
\cmidrule(lr){3-8}
\multirow{-2}{*}{\textbf{Methods}}
& \multirow{-2}{*}{\textbf{Settings}}
& BU & Atr & ER & RR & CI & Avg \\
\midrule
Qwen3-VL & $1/4$, Passive
& 68.9 & 17.8 & 45.1 & 50.4 & 64.7 & 49.4 \\
Qwen3-VL & $1/4$, Dynamic
& 75.6 & 22.3 & 46.2 & 53.4 & 71.0 & 53.7 \\
CamVLM & $1/4$, Dynamic
& \textbf{76.2}
& \textbf{28.3}
& \textbf{47.4}
& \textbf{66.0}
& \textbf{75.2}
& \textbf{58.8} \\
\midrule
Qwen3-VL & $1/6$, Passive
& 60.3 & 10.5 & 42.9 & 49.6 & 55.8 & 43.9 \\
Qwen3-VL & $1/6$, Dynamic
& 64.4 & 19.8 & 45.4 & 53.0 & 58.9 & 48.3 \\
CamVLM & $1/6$, Dynamic
& \textbf{67.6}
& \textbf{30.7}
& \textbf{47.8}
& \textbf{67.2}
& \textbf{76.9}
& \textbf{58.6} \\
\midrule
Qwen3-VL & $1/9$, Passive
& 57.1 & 10.5 & 41.2 & 46.2 & 51.3 & 41.3 \\
Qwen3-VL & $1/9$, Dynamic
& 67.2 & 17.2 & 44.5 & 51.3 & 51.3 & 45.9 \\
\rowcolor[RGB]{235,235,235}
CamVLM & $1/9$, Dynamic
& \textbf{70.6}
& \textbf{29.4}
& \textbf{46.2}
& \textbf{64.7}
& \textbf{72.7}
& \textbf{57.0} \\
\bottomrule
\end{tabular}%
}
\vspace{-0.15in}
\end{wraptable}

\paragraph{Effect of Different Viewpoint Sizes.}
We further evaluate different viewpoint sizes by training CamVLM from scratch with initial viewpoints covering the central $1/4$ and $1/6$ of the full frame, and comparing it with the corresponding passive and dynamic baselines. As shown in Table~\ref{tab:ablation_viewpoint_size}, CamVLM consistently outperforms the baselines across all settings. Larger viewpoints often capture the target immediately, whereas smaller viewpoints may initially miss it. Nevertheless, SFT enables the model to first zoom out for target search and then perform precise tracking, recovering complete fine-grained visual evidence.

\begin{wraptable}{r}{0.5\textwidth}
\vspace{-0.3in}
\caption{Ablation study of different training pipelines on UDVideoQA\(^*\). The best results are in bold. The default configuration is in gray.}
\label{tab:ablation_training}
\centering
\resizebox{\linewidth}{!}{%
\begin{tabular}{lc|cccccc}
\toprule
& & \multicolumn{6}{c}{\textbf{UDVideoQA\(^*\)}} \\
\cmidrule(lr){3-8}
\multirow{-2}{*}{\textbf{Methods}}
& \multirow{-2}{*}{\textbf{Settings}}
& BU & Atr & ER & RR & CI & Avg \\
\midrule
CamVLM (SFT)
& Dynamic
& 65.1 & 27.7 & 38.2 & 63.0 & 72.4 & 53.7 \\
CamVLM (RL)
& Dynamic
& \textbf{71.8}
& 25.2
& \textbf{54.2}
& 54.6
& \textbf{72.7}
& 56.0 \\
\rowcolor[RGB]{235,235,235}
CamVLM (SFT+RL)
& Dynamic
& 70.6
& \textbf{29.4}
& 46.2
& \textbf{64.7}
& \textbf{72.7}
& \textbf{57.0} \\
\bottomrule
\end{tabular}%
}
\vspace{-0.15in}
\end{wraptable}

\paragraph{Effect of Different Training Pipelines.}
Table~\ref{tab:ablation_training} compares different training pipelines. SFT alone already achieves strong performance, indicating that it effectively learns viewpoint control and surveillance video understanding. Applying RL directly to Qwen3-VL-8B-Instruct also brings substantial gains, showing that these capabilities can emerge through autonomous exploration. Combining SFT with RL achieves the best performance, demonstrating that SFT provides a strong initialization for subsequent RL refinement.

\begin{wrapfigure}{r}{0.5\textwidth}
\vspace{-0.2in}
\centering
\includegraphics[width=\linewidth]{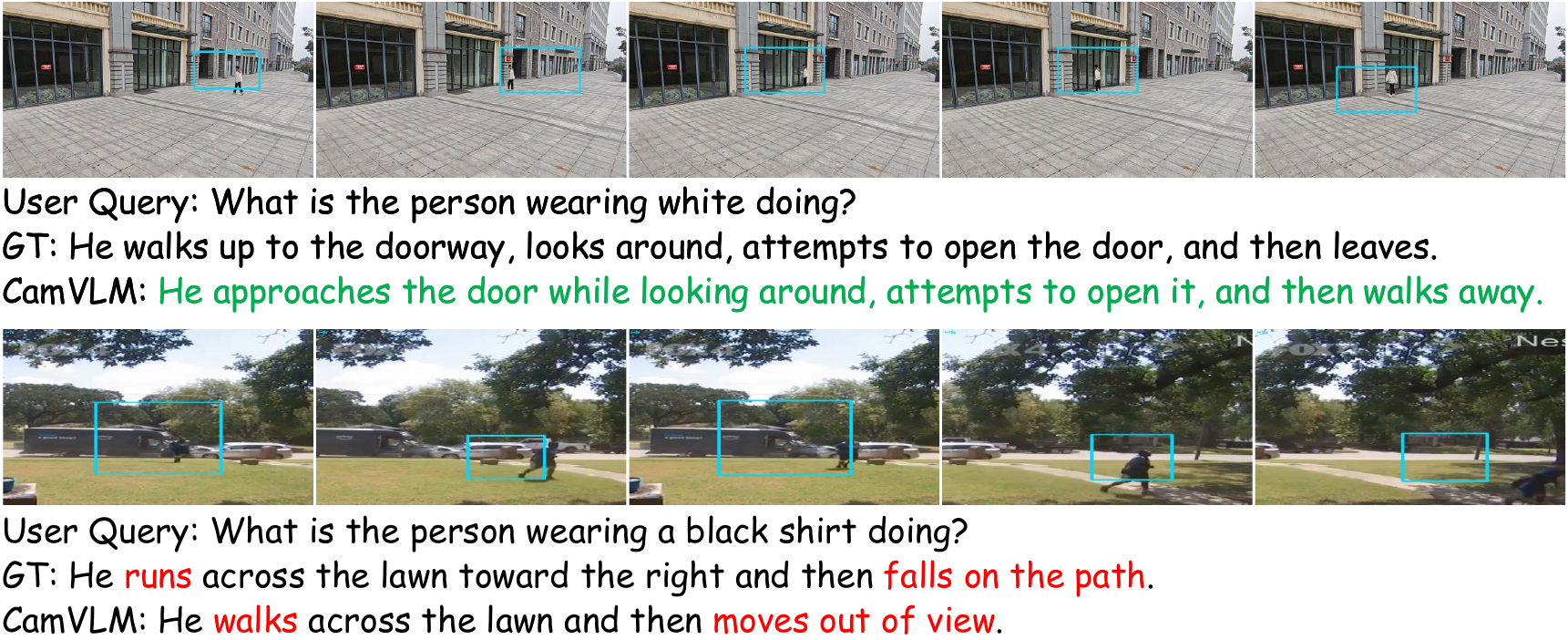}
\caption{Qualitative analysis of CamVLM on fine-grained human action understanding. CamVLM accurately recognizes complete action sequences in simple scenarios (top), but may miss subtle motion semantics and critical events in challenging videos (bottom), highlighting remaining challenges in fine-grained temporal reasoning.}
\label{fig:case_analysis}
\vspace{-0.25in}
\end{wrapfigure}

\subsection{Qualitative Analysis}

Figure~\ref{fig:case_analysis} provides qualitative examples of CamVLM. In simple interaction scenarios, CamVLM accurately captures complete action sequences by actively tracking the target through dynamic viewpoint control, maintaining informative visual evidence over time. In more challenging cases involving subtle motion differences and unexpected events, the model may still miss critical actions or confuse similar motions, such as describing running as walking and overlooking falls. These examples demonstrate that dynamic viewpoint control improves object-centric evidence acquisition and action understanding, while highlighting remaining challenges in fine-grained temporal reasoning.
\section{Conclusion}
\label{sec:conclusion}
In this work, we introduce CamVLM, an active visual reasoning framework that enables LVLMs to dynamically control camera viewpoints to actively acquire visual evidence in real-world surveillance. By constructing CCTV-Anomaly and CamTrack-53K, and formulating viewpoint control as a sequential decision-making problem, CamVLM learns effective camera policies. Extensive experiments demonstrate that active camera-based reasoning improves fine-grained understanding of real-world surveillance by acquiring more informative visual evidence.

\subsection*{AI Use Statement}
In this work, we used Large Language Models (LLMs) to improve the clarity, grammar, and readability of the manuscript. After using these tools, we carefully reviewed and edited the content as needed and take full responsibility for the final content of this work.

\subsection*{Ethics Statement}
This work studies video understanding in surveillance scenarios, where visual data may contain individuals and potentially sensitive events. Our research focuses on event-level understanding and active visual perception rather than personal identification. We acknowledge the potential privacy and misuse risks associated with surveillance technologies and encourage responsible use of the proposed datasets and models in accordance with applicable data licenses, privacy requirements, and ethical guidelines.

\subsection*{Reproducibility Statement}
To facilitate reproducibility, we provide detailed descriptions of the model architecture, training pipeline, viewpoint control mechanism, dataset construction, evaluation protocols, and implementation settings in the main paper and appendix. We also provide the prompt templates used for training and evaluation. The full datasets, models, and code will be released to support reproduction of our results and future research.

\bibliography{iclr2027_conference}
\bibliographystyle{iclr2027_conference}
\clearpage
\appendix
\startcontents[appendix]

\begin{center}
{\Large
\textbf{Appendix}
\par}
\end{center}

\vspace{1em}

\noindent{\Large\scshape Contents\par}

\vspace{0.5em}

\printcontents[appendix]{}{1}[2]{}

\clearpage

\section{Limitations}
\label{limitations}
\paragraph{Virtual-to-Physical Camera Control.}
Large-scale viewpoint-action trajectories from physical cameras operating in real-world, real-time surveillance are difficult to collect. We therefore adopt a virtual camera simulation, treating each full video frame as a panoramic observation space and a local region as the current camera viewpoint. Within this space, CamVLM learns camera control policies through directional and zooming actions with corresponding parameters, focusing on how and where the camera should move rather than device-specific actuation. For real-world deployment, the predicted actions and parameters could be translated into physical camera commands according to camera calibration and hardware specifications. Direct validation on physical camera systems remains an important direction for future work.

Current viewpoint trajectories still rely on predefined rules and fixed adjustment intervals, which limits adaptability in complex scenarios. Moreover, the framework focuses on single-camera control, while real-world surveillance requires more autonomous and collaborative perception. Future work will explore physical camera deployment, adaptive trajectory generation, dynamic viewpoint scheduling, and multi-camera coordination for more intelligent visual reasoning systems.

\section{Evaluation Details on UDVideoQA$^{*}$}
Since the original test split of UDVideoQA~\citep{vishal2026udvideoqa} is no longer available, we construct UDVideoQA$^{*}$ with permission and approval from the original authors using Set~03 and Set~20 from the released dataset, as these two subsets provide the greatest question diversity. The resulting benchmark contains five question types: Basic Understanding (BU), Attribution (Atr), Event Reasoning (ER), Reverse Reasoning (RR), and Counterfactual Inference (CI). Each question type contains 238 QA pairs, resulting in a total of 1,190 QA pairs.

Following the standard evaluation protocol of UDVideoQA, we employ an external large language model (LLM) as a semantic matching evaluator. Specifically, we use GPT-5.5~\citep{openai2026gpt55} as a text-only evaluator. To mitigate potential evaluation bias, all model judgments are further verified by human annotators, who independently assess the correctness of the generated answers. Each response is assigned a binary score $S_{\mathrm{LLM},i} \in \{0,1\}$, indicating whether the predicted answer is semantically consistent with the ground-truth answer.

We further adopt the weighted scoring scheme introduced by UDVideoQA to account for differences in cognitive complexity across reasoning categories. More challenging reasoning types are assigned larger weights: 1.0 for BU, 1.2 for Atr, 1.3 for ER and RR, and 1.5 for CI. The final weighted score $W$ is computed as
\begin{equation}
W = \frac{1}{N}\sum_{i=1}^{N} S_{\mathrm{LLM},i} \cdot W_{c_i},
\end{equation}
where $S_{\mathrm{LLM},i}$ denotes the correctness score of the $i$-th response, and $W_{c_i}$ is the weight assigned to its reasoning category $c_i$. This formulation places greater emphasis on successful predictions for cognitively demanding question types, particularly counterfactual inference, which is designed to reduce hallucinations and encourage deeper causal understanding. We report the accuracy for each of the five question types together with the weighted overall accuracy.

For all evaluations, we set the temperature to 0 and adopt greedy decoding to ensure deterministic outputs. All reported results are obtained from a single evaluation run using the final trained model checkpoint.

\section{Details of CCTV-Anomaly Construction}
\label{sec:sup_cctv}
\subsection{Video Collection}
CCTV-Anomaly is constructed from two complementary sources: publicly available surveillance anomaly datasets and newly collected real-world surveillance videos. To ensure data quality and suitability for LVLM training, we perform extensive data filtering to remove duplicate videos, manually edited clips, severely occluded scenes, and samples with ambiguous event descriptions. After filtering, the public subset contains 5,832 videos collected from existing datasets, including XD-Violence~\citep{wu2020not}, CCTV-Fights~\citep{perez2019detection}, GTA-Crime~\citep{kim2025gta}, and UBnormal~\citep{acsintoae2022ubnormal}. The newly collected subset contains 8,301 videos obtained from diverse real-world surveillance resources, covering various environments, viewpoints, and event scenarios.

\subsection{Multi-level Video Annotation}
To support fine-grained surveillance video understanding, we develop a multi-level annotation pipeline consisting of event category annotation, detailed caption annotation, and structured evaluation annotation. The annotation process adopts an LLM-assisted strategy with human verification to ensure annotation reliability. We use Doubao-Seed-2.0-Pro~\citep{seed2026seed2} as the automatic annotator.

\paragraph{Event Category Annotation.}
Each video is assigned one event category according to our predefined taxonomy. For public datasets with existing labels, we inherit the original annotations when they are semantically consistent with our categories. For unlabeled public videos and newly collected samples, event categories are generated through video analysis and subsequently verified by human annotators.

\paragraph{Event Caption Annotation.}
Each video is further annotated with a concise yet informative caption describing the key semantic information of the event, including involved subjects, major actions, and event outcomes. Instead of exhaustively describing all visual contents, the captions focus on information relevant to surveillance reasoning. We adopt an LLM-assisted caption generation pipeline, followed by human inspection to correct inaccurate descriptions and ensure consistency with the corresponding event category.

\paragraph{Structured Evaluation Annotation.}
To enable fine-grained evaluation of LVLM outputs, we randomly select 1,000 videos as the test set and construct structured semantic annotations from verified captions. Specifically, each video is annotated with five semantic components: location, time of day, subjects, activities, and objects. All structured annotations are manually reviewed to ensure consistency and completeness.

\section{Details of CamTrack-53K Construction}
\label{sec:camtrack_construction}

\subsection{Data Curation}
We construct CamTrack-53K from the A2D~\citep{xu2015can}, 
MeViS~\citep{ding2023mevis}, and Refer-YouTube-VOS~\citep{seo2020urvos} 
subsets of VideoRefer-700K~\citep{yuan2025videorefer}. The resulting 
dataset contains 53,059 samples, including 26,221 from A2D, 
8,459 from MeViS, and 18,379 from Refer-YouTube-VOS.

\subsection{Viewpoint Representation}
The viewpoint is represented as a normalized rectangular region:
\begin{equation}
v_t =
\left[
x_{1,t},
y_{1,t},
x_{2,t},
y_{2,t}
\right],
\end{equation}
where $t$ denotes the viewpoint step, and $(x_{1,t}, y_{1,t})$ and $(x_{2,t}, y_{2,t})$ denote the top-left and bottom-right coordinates of the viewpoint at step $t$, respectively. All coordinates are normalized to $[0,1]$.

For our main experiments, the viewpoint is initialized at the center
of the frame and covers $1/9$ of the full-frame area:
\begin{equation}
v_0 =
\left[
\frac{1}{3},
\frac{1}{3},
\frac{2}{3},
\frac{2}{3}
\right].
\end{equation}
We additionally construct variants with initial viewpoint area ratios of $1/6$ and $1/4$ for ablation studies.

\subsection{Viewpoint Construction}
Each clip is associated with a viewpoint action derived from the
spatial configuration of the task-relevant objects. For clips
involving multiple objects, their bounding boxes are merged into the
minimum enclosing box:
\begin{equation}
b_t =
\left[
\min_i x_{1,t}^{(i)},
\min_i y_{1,t}^{(i)},
\max_i x_{2,t}^{(i)},
\max_i y_{2,t}^{(i)}
\right].
\end{equation}
Here, $t$ denotes the viewpoint step, and $i$ indexes the
task-relevant objects. $(x_{1,t}^{(i)}, y_{1,t}^{(i)})$ and
$(x_{2,t}^{(i)}, y_{2,t}^{(i)})$ denote the top-left and bottom-right
coordinates of the bounding box of the $i$-th object, respectively.
The resulting $b_t$ represents the aggregated target region at step
$t$.

The target viewpoint is designed to fully cover the task-relevant objects
while preserving sufficient surrounding context. Let $W$ and $H$
denote the width and height of the video frame, respectively, and let
$w(b_t)$, $h(b_t)$, and $A(b_t)$ denote the width, height, and area of
the target bounding box. We adaptively determine the normalized side
length of the viewpoint according to the size of the target box
relative to the full frame:
\begin{equation}
s_{v_t}
=
\operatorname{clip}
\left(
\max\left\{
\sqrt{\frac{A(b_t)}{\rho WH}},
\frac{\kappa w(b_t)}{W},
\frac{\kappa h(b_t)}{H}
\right\},
s_{\min},
s_{\max}
\right),
\end{equation}
where $\rho=0.2$ controls the desired proportion of the viewpoint
occupied by the target, and $\kappa=1.3$ is a padding factor that
preserves visual context around the target. The parameters $s_{\min}=0.2$
and $s_{\max}=0.8$ constrain the minimum and maximum viewpoint scales,
respectively. Consequently, smaller targets are associated with more
compact viewpoints, whereas larger targets require broader
viewpoints.

The target viewpoint is centered on the target bounding box and
appropriately shifted when approaching the frame boundaries, ensuring
that it always remains within the valid image region. When the target
is already fully covered, approximately centered, and observed at a
suitable scale, the current viewpoint is maintained to avoid
unnecessary viewpoint adjustments.

\subsection{Viewpoint Trajectory Visualization}
\begin{figure*}[ht]
\centering
\includegraphics[width=\textwidth]{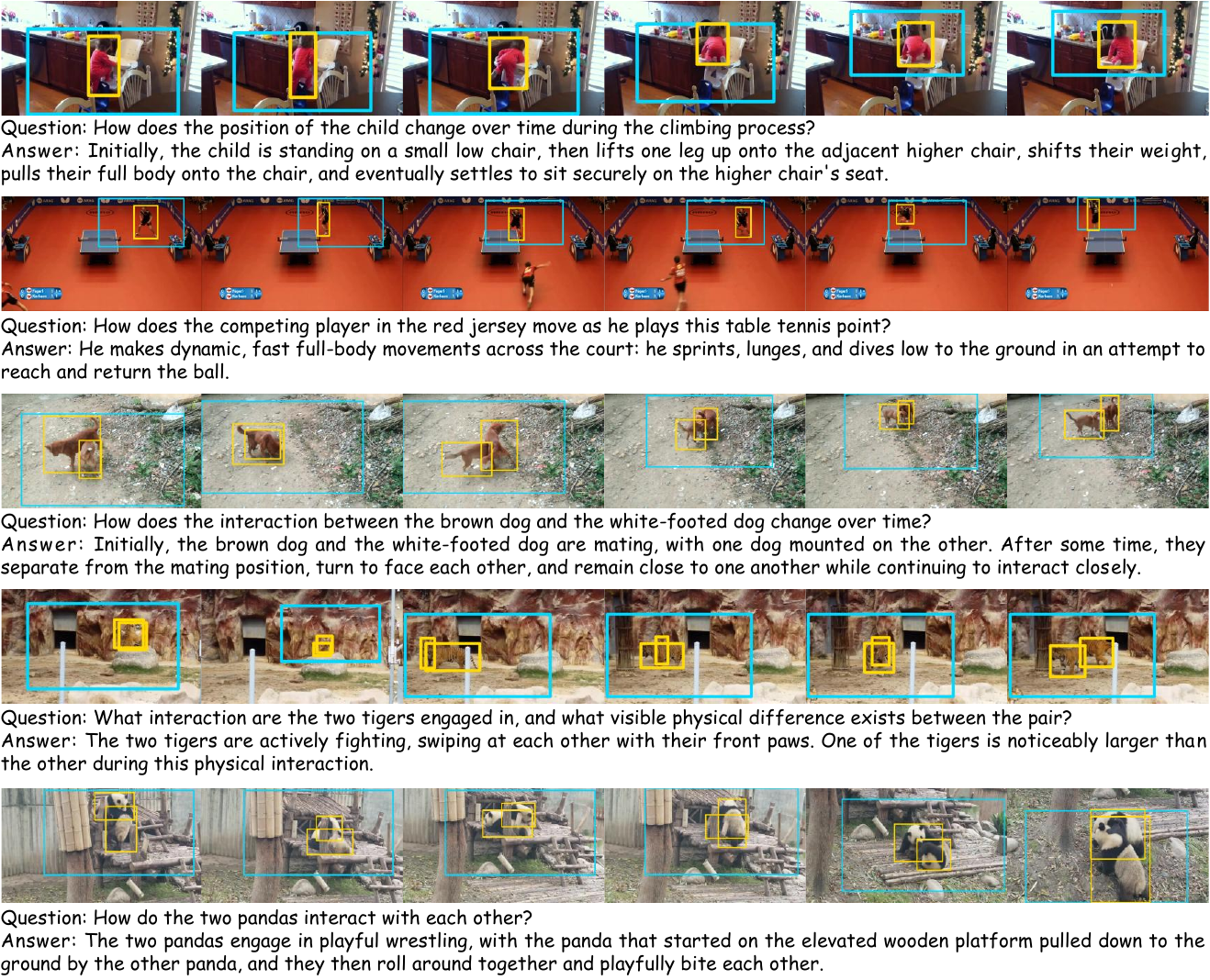}
\caption{Visualization of viewpoint trajectories in CamTrack-53K. Yellow and blue bounding boxes denote object bounding boxes and current viewpoints, respectively.}
\label{fig:vis}
\end{figure*}

We visualize representative temporally ordered viewpoint trajectories from CamTrack-53K in Figure~\ref{fig:vis}, where the yellow boxes denote object bounding boxes and the blue boxes indicate the current viewpoints. As shown, the constructed viewpoints continuously and stably cover task-relevant objects throughout their motion, while dynamically adjusting their positions and spatial extents according to object scale and spatial distribution. Meanwhile, they suppress excessive background content while preserving sufficient surrounding context and interaction cues, thereby yielding compact, coherent, and informative fine-grained visual evidence.

\section{Prompt Template}
\begin{figure*}[t]
\centering
\includegraphics[width=\textwidth]{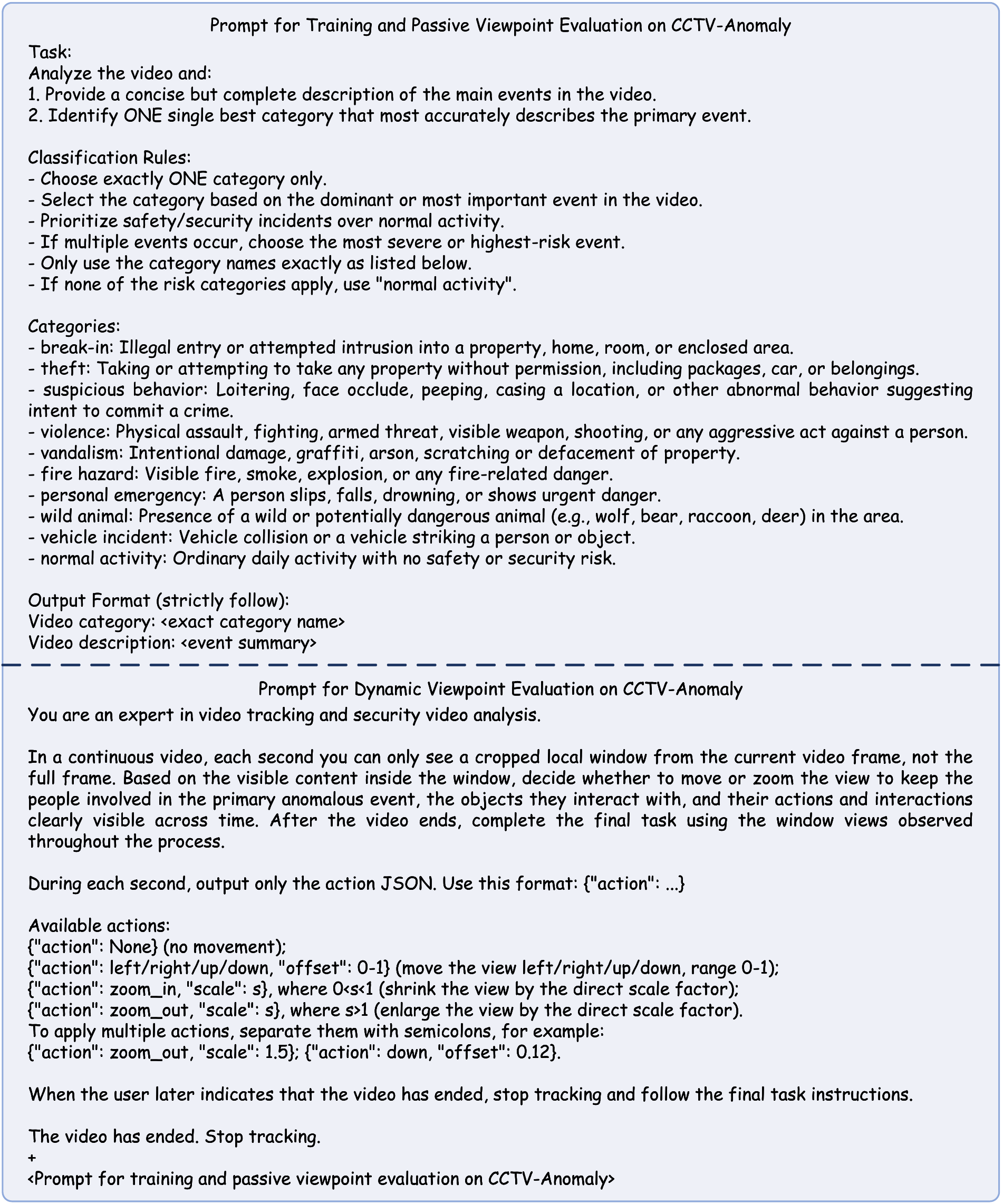}
\caption{Prompt templates for training and passive and dynamic viewpoint evaluation on CCTV-Anomaly.}
\label{fig:prompt_1}
\end{figure*}

\paragraph{Prompt Template for Training and Passive Viewpoint Evaluation on CCTV-Anomaly.}
Figure~\ref{fig:prompt_1} (top) illustrates the prompt template for training and passive viewpoint evaluation on CCTV-Anomaly.

\paragraph{Prompt Template for Dynamic Viewpoint Evaluation on CCTV-Anomaly.}
Figure~\ref{fig:prompt_1} (bottom) illustrates the prompt template for dynamic viewpoint evaluation on CCTV-Anomaly.

\begin{figure*}[t]
\centering
\includegraphics[width=\textwidth]{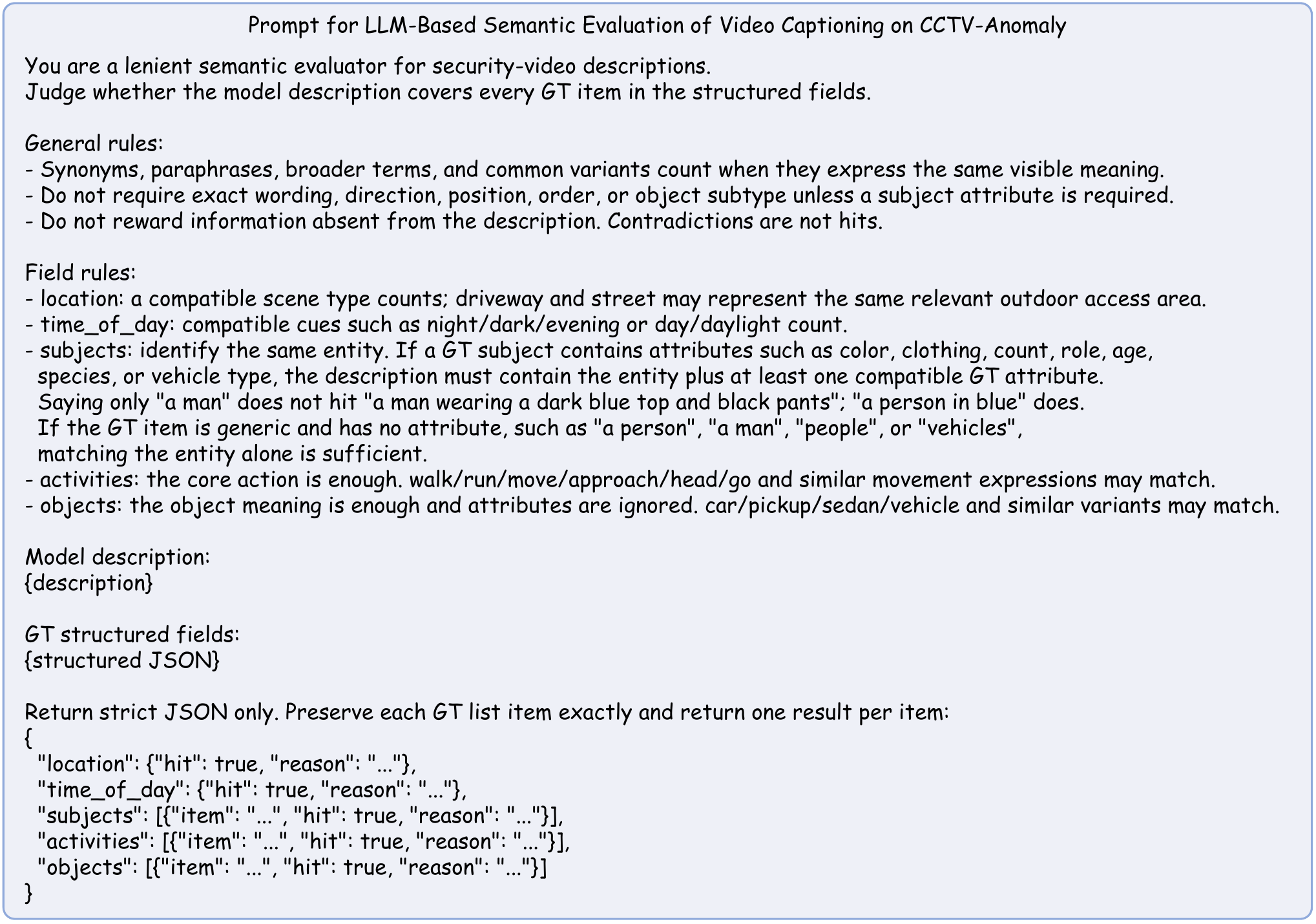}
\caption{Prompt template for LLM-based semantic evaluation of video captioning on CCTV-Anomaly.}
\label{fig:prompt_2}
\end{figure*}

\paragraph{Prompt Template for LLM-Based Semantic Evaluation of Video Captioning on CCTV-Anomaly.}
Figure~\ref{fig:prompt_2} illustrates the prompt template for LLM-based semantic evaluation of video captioning on CCTV-Anomaly.

\begin{figure*}[t]
\centering
\includegraphics[width=\textwidth]{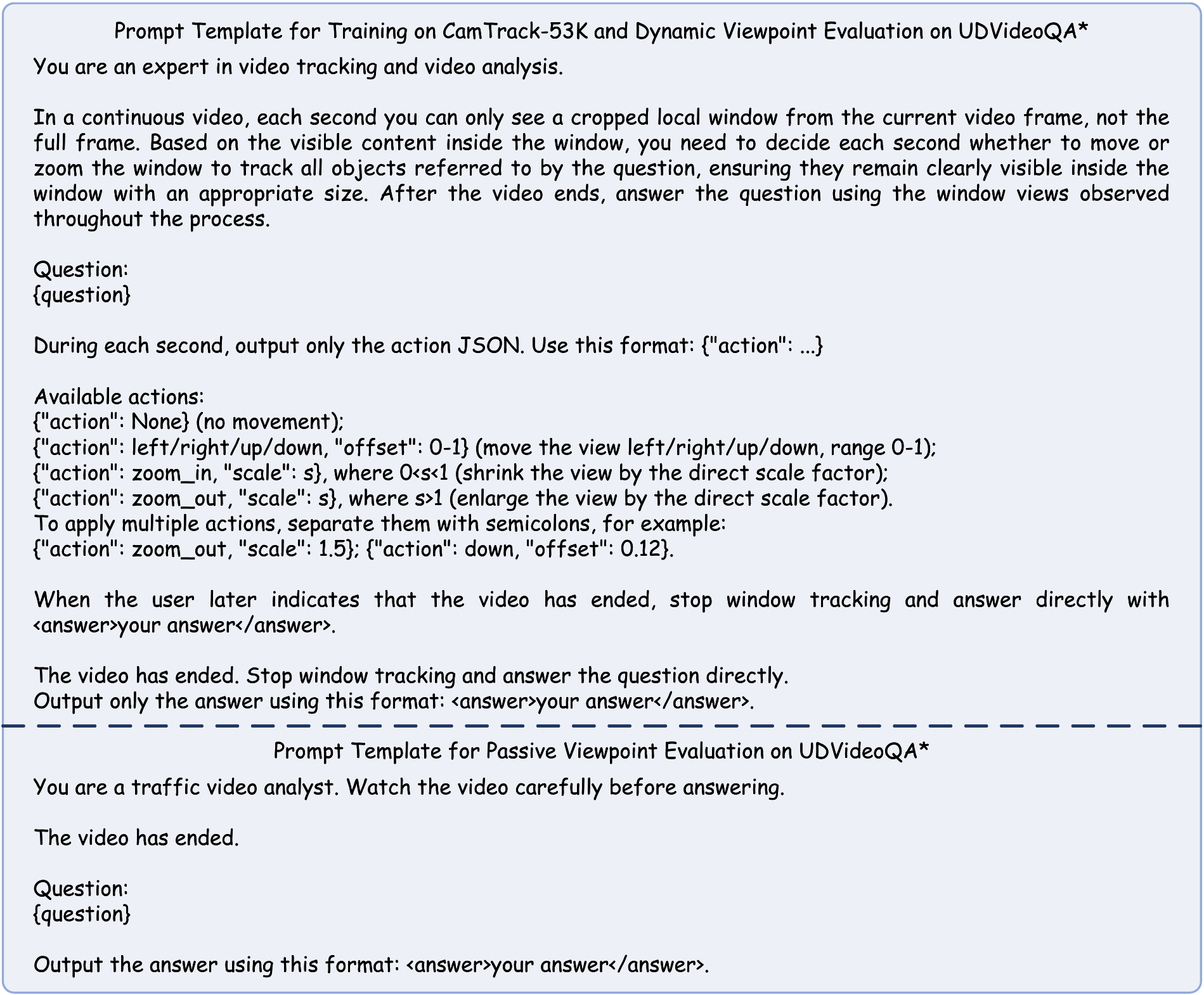}
\caption{Prompt templates for training on CamTrack-53K and dynamic and passive viewpoint evaluation on UDVideoQA\(^*\).}
\label{fig:prompt_3}
\end{figure*}

\paragraph{Prompt Template for Training on CamTrack-53K and Dynamic Viewpoint Evaluation on UDVideoQA\(^*\).}
Figure~\ref{fig:prompt_3} (top) illustrates the prompt template for training on CamTrack-53K and dynamic viewpoint evaluation on UDVideoQA\(^*\).

\paragraph{Prompt Template for Passive Viewpoint Evaluation on UDVideoQA\(^*\).}
Figure~\ref{fig:prompt_3} (bottom) illustrates the prompt template for passive viewpoint evaluation on UDVideoQA\(^*\).

\begin{figure*}[t]
\centering
\includegraphics[width=\textwidth]{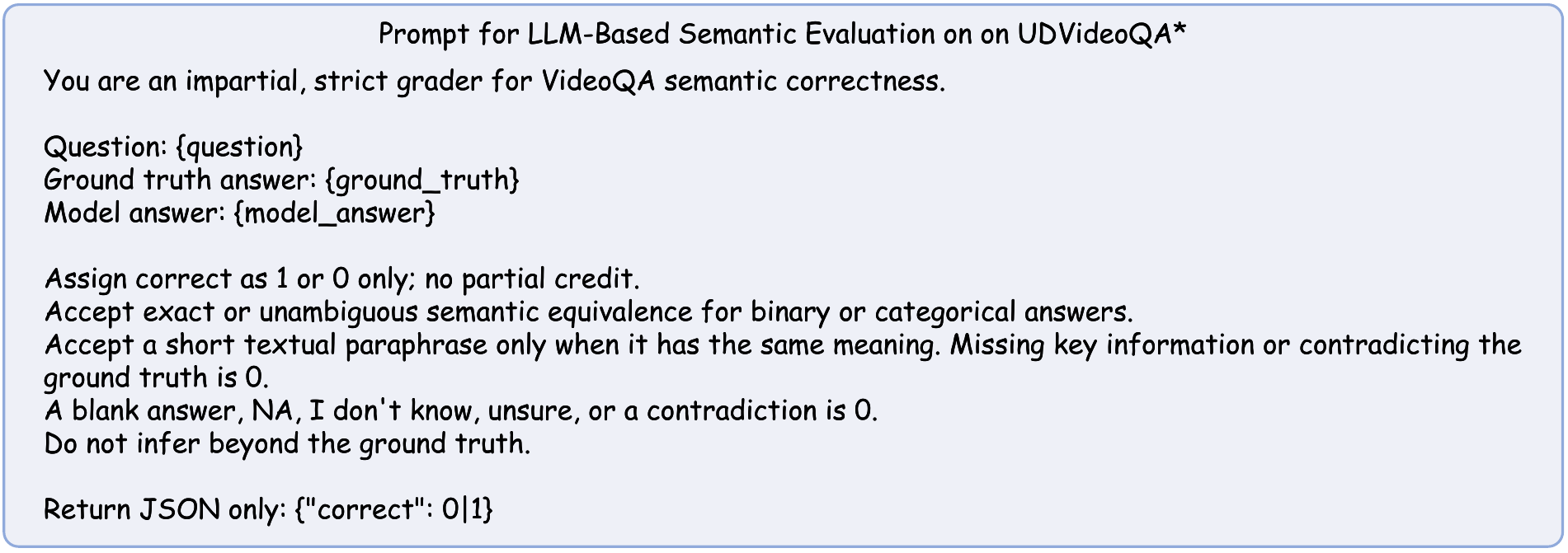}
\caption{Prompt template for LLM-based semantic evaluation on UDVideoQA\(^*\).}
\label{fig:prompt_4}
\end{figure*}

\paragraph{Prompt Template for LLM-Based Semantic Evaluation on UDVideoQA\(^*\).}
Figure~\ref{fig:prompt_4} illustrates the prompt template for LLM-based semantic evaluation on UDVideoQA\(^*\).

\clearpage

\end{document}